\documentclass[11pt,a4paper]{article}
\usepackage[utf8]{inputenc}
\usepackage[T1]{fontenc}
\usepackage{amsmath}
\usepackage{amssymb}
\usepackage{graphicx}
\usepackage{booktabs}
\usepackage{hyperref}
\usepackage{geometry}
\title{Understanding Context Sampling in TabPFN\\ on Small Tabular Datasets}

\author{
  Mohammed Abdullah\\
  Department of AI \& Data Science\\
  Anna University\\
  \texttt{2023176026@student.annauniv.edu}
}

\date{\today}

\begin{document}

\maketitle

\begin{abstract}
TabPFN classifies by in-context learning: it conditions on a set of labeled
training rows (the context, or prototypes) and predicts test labels without
gradient updates. On small tabular datasets, where TabPFN is designed to be
used, the practitioner still chooses how large that context is and which rows
go into it. We ask how much those choices matter. Using repeated random
sub-sampling on 15 small OpenML datasets, we study three linked questions:
(i) how variable predictions are across different random contexts of the same
size, and whether larger contexts are more stable; (ii) whether accuracy tracks
how well a sampled context preserves the full training distribution; and
(iii) whether more expensive selection (K-Means, farthest-point sampling) buys
anything over uniform random selection once the context is representative.
We find that larger contexts are both more accurate and markedly more stable
(the coefficient of variation of AUC falls from roughly 6--18\% at $k{=}16$ to
1--4\% at the largest $k$ on the datasets with room to improve). Accuracy does
correlate with how well a random context preserves the training distribution,
but a controlled experiment overturns the causal reading: when we construct
contexts that match the feature means closely, accuracy drops substantially (by
up to 0.5 AUC), because that construction also destroys diversity. Controlled
experiments and a mixed-effects analysis point to diversity and coverage, rather
than feature-mean matching, as the factor that predicts accuracy (diversity
$\beta=+0.23$, $p=3\times10^{-12}$; feature-mean shift $\beta=-0.01$, $p=0.71$
once diversity is included). This also explains why K-Means and farthest-point
selection, which are diverse but not distribution-preserving, match uniform
random while costing two to three orders of magnitude more. Random sampling works because it
covers the feature space in expectation, not because it matches its
distribution. We report effect sizes and confidence intervals throughout.

\end{abstract}

\section{Introduction}

Prototype selection is a fundamental inference-time decision in TabPFN
\cite{hollmann2025tabpfn}, a prior-data fitted network \cite{muller2022pfn}: the
model conditions on a chosen set of labeled rows (the context) and predicts in a
single forward pass, the tabular instance of the in-context learning paradigm
popularized by large language models \cite{brown2020gpt3, dong2024icl}. Prototype and instance selection have a long
history in nearest-neighbor learning \cite{garcia2012prototype}, and
sophisticated selection methods (clustering, coverage heuristics) have been
proposed; yet it remains unclear whether their gains come from the selection
\emph{algorithm} or simply from selecting a subset that is representative of the
training data. We investigate this on small tabular
datasets, the regime TabPFN targets, where we can afford many repeated draws
and isolate the effect of the context from confounds of scale.

We study three connected questions, on 15 small OpenML datasets:

\begin{description}
\item[H1 (Stability).] Different random contexts of the same size $k$ yield
different predictions; the spread of accuracy across random draws shrinks as $k$
grows.
\item[H2 (What makes a context good).] We ask whether accuracy is driven by how
well the context preserves the training distribution (representativeness) or by
how well it covers the feature space (diversity). A correlation initially
suggests representativeness; a controlled experiment shows the causal factor is
diversity.
\item[H3 (Selection cost).] Expensive selection methods (K-Means, FPS) add
little accuracy over uniform random selection despite far higher cost.
\end{description}

These are mechanistic questions about \emph{how} TabPFN responds to its context,
and together they form a single account: context size controls reliability (H1),
diversity and coverage (not distribution matching) explain which contexts succeed
(H2), and this explains why cheap random sampling, which covers the space in
expectation, is hard to beat (H3). We answer them with repeated random
sub-sampling, \emph{controlled} construction of contexts with targeted geometry,
and two supporting stress tests (duplicating and dropping part of the context).
Our contribution is to identify context diversity, rather than representativeness
or the choice of selection algorithm, as what a small-data context needs, and to
reach that conclusion through a controlled experiment that overturns the more
obvious correlational story.

\section{Related Work}

\textbf{In-context learning and TabPFN.} In-context learning, in which a model
adapts to a task from examples placed in its input without weight updates, was
popularized by large language models \cite{brown2020gpt3} and has since been
surveyed extensively \cite{dong2024icl}. Prior-data fitted networks (PFNs) bring
this paradigm to supervised learning by meta-training a transformer to perform
Bayesian inference in a forward pass \cite{muller2022pfn}; TabPFN applies it to
tabular classification and regression \cite{hollmann2025tabpfn}. Our study takes
TabPFN as given and asks a downstream question its design raises but does not
answer: how the in-context set should be chosen on small data.

\textbf{Prototype and instance selection.} Choosing a small subset of training
examples is a classical problem in nearest-neighbor learning, where prototype
selection methods aim to shrink the reference set while preserving accuracy
\cite{garcia2012prototype}. Clustering- and coverage-based selectors are standard
tools: k-means++ seeding \cite{arthur2007kmeans} and farthest-point / $k$-center
sampling \cite{gonzalez1985clustering} both spread selected points across the
input space. We evaluate exactly these families as context selectors for TabPFN
and find that, at matched budget, they do not beat uniform random enough to
justify their cost.

\textbf{Coverage, diversity, and coresets.} A parallel line of work formalizes
``covering'' a dataset with a small set. Determinantal point processes model
diverse subsets through a volume (log-determinant) objective
\cite{kulesza2012dpp}; coreset constructions seek small weighted subsets that
approximate a learning objective \cite{feldman2020coresets}, and coverage-driven
selection has been used directly for active learning \cite{sener2018coreset}.
These motivate the geometric diversity/coverage descriptors we use, and our
controlled experiments provide evidence that, for TabPFN on small data, such
coverage rather than distributional fidelity is what tracks accuracy.

\textbf{Data valuation and compression.} Related efforts ask which data are
valuable or how to compress a dataset: data-valuation methods score individual
examples by their marginal contribution \cite{ghorbani2019datashapley}, and
dataset distillation synthesizes tiny training sets that reproduce full-data
performance \cite{wang2018distillation}. Our aim is narrower and diagnostic:
rather than propose a new selection or synthesis method, we isolate \emph{which
property of a context} drives TabPFN's accuracy, using controlled construction
and mixed-effects inference \cite{bates2015lme4}.

\textbf{Benchmark data.} All experiments use public tasks from OpenML
\cite{vanschoren2014openml}, which supplies the standardized, citable datasets
that make the study reproducible.

\section{Setup}

We use TabPFN v3 with \texttt{ignore\_pretraining\_limits=True}. Each dataset is
split 80/20 (stratified, fixed split seed). A context of size $k$ is formed by
selecting $k$ training rows and fitting TabPFN on exactly those rows
(\texttt{clf.fit(X[idx], y[idx])}); the held-out 20\% is the test set throughout.
We report ROC AUC (one-vs-rest macro for the multiclass dataset). For the
stability and correlational analyses we draw \textbf{20 independent random
contexts} per $k$. For the causal analyses in H2 we additionally \emph{construct}
contexts with targeted geometry (Section on the controlled test) rather than
sampling them; for selection-method and stress tests we use three or more seeds.

\begin{table}[h]
\centering
\caption{Datasets (all small; public OpenML \cite{vanschoren2014openml}).}
\label{tab:datasets}
\begin{tabular}{lrrr}
\toprule
Dataset & rows & features & classes \\
\midrule
diabetes           & 768  & 8  & 2 \\
credit-g           & 1000 & 20 & 2 \\
vehicle            & 846  & 18 & 4 \\
wdbc               & 569  & 30 & 2 \\
breast-w           & 699  & 9  & 2 \\
blood-transfusion  & 748  & 4  & 2 \\
ionosphere         & 351  & 34 & 2 \\
sonar              & 208  & 60 & 2 \\
glass              & 214  & 9  & 6 \\
australian         & 690  & 14 & 2 \\
heart-statlog      & 270  & 13 & 2 \\
segment            & 2310 & 18 & 7 \\
waveform-5000      & 5000 & 40 & 3 \\
spambase           & 4601 & 57 & 2 \\
kc1                & 2109 & 21 & 2 \\
\bottomrule
\end{tabular}
\end{table}

\textbf{Context descriptors.} Let the full training set have standardized
feature matrix $X\in\mathbb{R}^{n\times m}$ with column means $\bar{X}$, and let
a context $C$ of size $k$ have feature rows $\{x_i\}_{i\in C}$ with mean
$\bar{X}_C$. We characterize $C$ with two families of measures.

\emph{Representativeness} (how close to the full distribution): the
total-variation distance between the context's class distribution $p_C$ and the
full class distribution $p$, and the mean absolute per-feature mean shift,
\begin{equation}
\texttt{class\_tv} = \tfrac{1}{2}\sum_{c} \lvert p_c - p_{C,c}\rvert,
\qquad
\texttt{feat\_shift} = \frac{1}{m}\sum_{j=1}^{m}
\bigl\lvert \bar{X}_{C,j} - \bar{X}_{j}\bigr\rvert .
\end{equation}
Smaller values mean $C$ looks more like the whole training set.

\emph{Diversity / coverage} (how much of the space $C$ spans): the mean pairwise
Euclidean distance among context rows and the log-determinant of the context
covariance $\Sigma_C$ (a volume),
\begin{equation}
\texttt{diversity} = \frac{1}{\lvert C\rvert^2}\sum_{i,i'\in C}
\lVert x_i - x_{i'}\rVert_2,
\qquad
\texttt{logdet\_cov} = \log\det\!\bigl(\Sigma_C + \varepsilon I\bigr).
\end{equation}
These are standard geometric coverage measures. Mean pairwise distance
summarizes spread, and the covariance log-determinant is the standard volume of
the region a point set occupies (it appears, for example, in determinantal point
processes \cite{kulesza2012dpp} and coreset constructions
\cite{feldman2020coresets, sener2018coreset}). The two families can vary independently, and
separating them is the crux of H2.

\section{H1: Context Stability}

\begin{figure}[h]
\centering
\includegraphics[width=\textwidth]{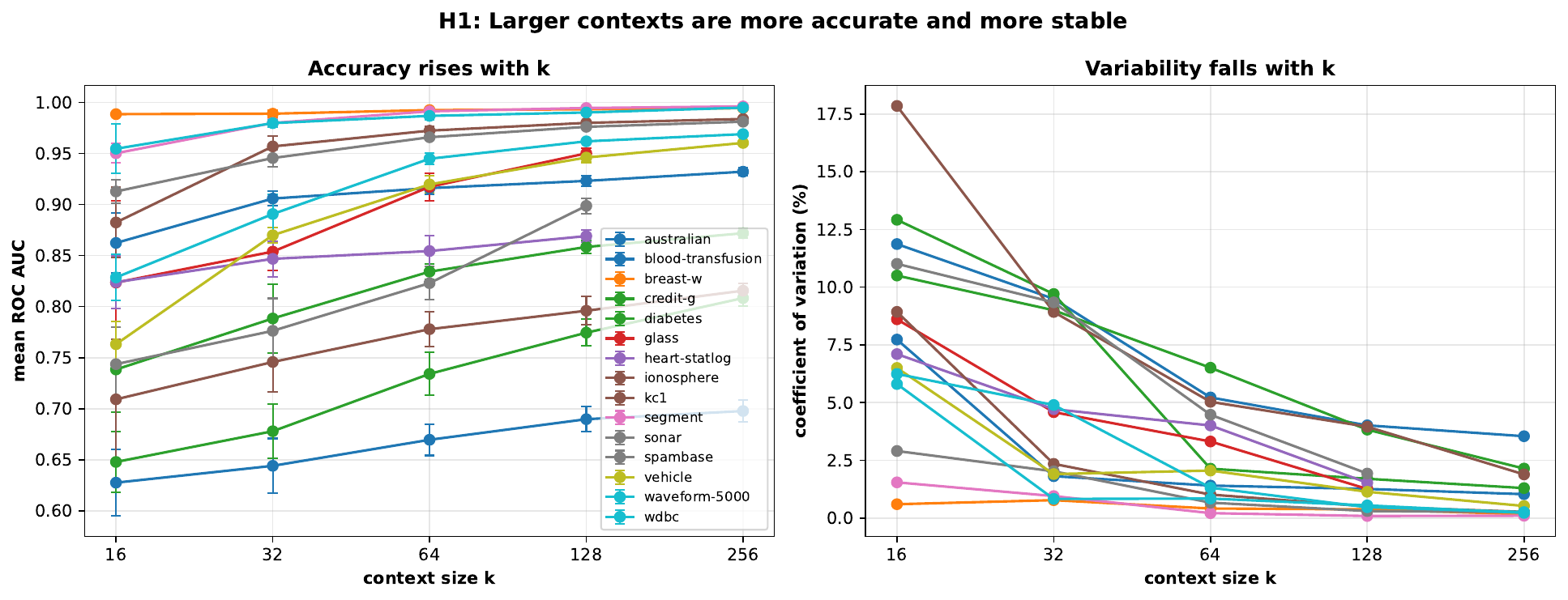}
\caption{Mean AUC (left) rises with context size $k$; the coefficient of
variation across 20 random draws (right) falls with $k$ on every dataset.}
\label{fig:stability}
\end{figure}

For each $k$ we drew 20 independent random contexts and recorded the resulting
test AUC. Table~\ref{tab:stability} reports the mean $\pm$ 95\% CI, and
Table~\ref{tab:stability_cv} the coefficient of variation
$\mathrm{CV}=\sigma/\mu$, a scale-free measure of variability, as a percentage.
Figure~\ref{fig:stability} plots mean AUC and CV against $k$.

\begin{table}[h]
\centering
\caption{Mean ROC AUC $\pm$ 95\% CI across 20 random contexts per $k$.
Dashes: $k{=}256$ exceeds the training-set size on the smallest datasets.}
\label{tab:stability}
\begin{tabular}{lrrrrr}
\toprule
Dataset & $k{=}16$ & $k{=}32$ & $k{=}64$ & $k{=}128$ & $k{=}256$ \\
\midrule
diabetes           & $.738{\pm}.042$ & $.788{\pm}.034$ & $.834{\pm}.008$ & $.858{\pm}.006$ & $.872{\pm}.005$ \\
credit-g           & $.648{\pm}.030$ & $.678{\pm}.027$ & $.734{\pm}.021$ & $.774{\pm}.013$ & $.808{\pm}.008$ \\
vehicle            & $.763{\pm}.022$ & $.870{\pm}.007$ & $.920{\pm}.008$ & $.946{\pm}.005$ & $.960{\pm}.002$ \\
wdbc               & $.955{\pm}.024$ & $.980{\pm}.004$ & $.987{\pm}.004$ & $.990{\pm}.002$ & $.995{\pm}.001$ \\
breast-w           & $.988{\pm}.003$ & $.989{\pm}.003$ & $.992{\pm}.002$ & $.993{\pm}.002$ & $.994{\pm}.001$ \\
blood-transfusion  & $.628{\pm}.033$ & $.644{\pm}.027$ & $.670{\pm}.015$ & $.690{\pm}.012$ & $.698{\pm}.011$ \\
ionosphere         & $.882{\pm}.035$ & $.957{\pm}.010$ & $.972{\pm}.004$ & $.980{\pm}.002$ & $.984{\pm}.001$ \\
sonar              & $.744{\pm}.036$ & $.776{\pm}.032$ & $.823{\pm}.016$ & $.899{\pm}.008$ & --- \\
glass              & $.824{\pm}.080$ & $.854{\pm}.018$ & $.917{\pm}.013$ & $.950{\pm}.005$ & --- \\
australian         & $.862{\pm}.029$ & $.906{\pm}.007$ & $.916{\pm}.006$ & $.923{\pm}.005$ & $.932{\pm}.004$ \\
heart-statlog      & $.824{\pm}.026$ & $.847{\pm}.018$ & $.854{\pm}.015$ & $.869{\pm}.006$ & --- \\
segment            & $.950{\pm}.010$ & $.980{\pm}.004$ & $.991{\pm}.001$ & $.994{\pm}.001$ & $.996{\pm}.001$ \\
waveform-5000      & $.829{\pm}.023$ & $.891{\pm}.019$ & $.945{\pm}.005$ & $.962{\pm}.002$ & $.969{\pm}.001$ \\
spambase           & $.913{\pm}.012$ & $.945{\pm}.008$ & $.966{\pm}.003$ & $.976{\pm}.001$ & $.981{\pm}.001$ \\
kc1                & $.709{\pm}.058$ & $.746{\pm}.029$ & $.778{\pm}.017$ & $.796{\pm}.014$ & $.816{\pm}.007$ \\
\bottomrule
\end{tabular}
\end{table}

\begin{table}[h]
\centering
\caption{Coefficient of variation ($\mathrm{CV}=\sigma/\mu$, \%) of ROC AUC
across the same 20 random contexts. CV falls as $k$ grows on every dataset.}
\label{tab:stability_cv}
\begin{tabular}{lrrrrr}
\toprule
Dataset & $k{=}16$ & $k{=}32$ & $k{=}64$ & $k{=}128$ & $k{=}256$ \\
\midrule
diabetes           & 12.9 & 9.7 & 2.1 & 1.7 & 1.3 \\
credit-g           & 10.5 & 9.0 & 6.5 & 3.8 & 2.1 \\
vehicle            & 6.5 & 1.9 & 2.1 & 1.1 & 0.5 \\
wdbc               & 5.8 & 0.8 & 0.8 & 0.6 & 0.2 \\
breast-w           & 0.6 & 0.8 & 0.4 & 0.4 & 0.2 \\
blood-transfusion  & 11.9 & 9.5 & 5.2 & 4.0 & 3.5 \\
ionosphere         & 8.9 & 2.3 & 1.0 & 0.5 & 0.3 \\
sonar              & 11.0 & 9.3 & 4.5 & 1.9 & --- \\
glass              & 8.6 & 4.6 & 3.3 & 1.3 & --- \\
australian         & 7.7 & 1.8 & 1.4 & 1.3 & 1.0 \\
heart-statlog      & 7.1 & 4.7 & 4.0 & 1.5 & --- \\
segment            & 1.5 & 1.0 & 0.2 & 0.1 & 0.1 \\
waveform-5000      & 6.2 & 4.9 & 1.3 & 0.5 & 0.3 \\
spambase           & 2.9 & 2.0 & 0.7 & 0.3 & 0.3 \\
kc1                & 17.8 & 8.9 & 5.0 & 4.0 & 1.9 \\
\bottomrule
\end{tabular}
\end{table}

\textbf{What we observe.} On every one of the 15 datasets, variability across
random contexts falls as $k$ grows. On datasets with accuracy headroom
(diabetes, credit-g, sonar, kc1, waveform, and others) the CV drops from roughly
6--18\% at $k{=}16$ to about 1--4\% at the largest $k$; datasets already near
ceiling (breast-w, segment) start low and still shrink. Mean AUC rises with $k$
in parallel. So on small data the context size is not only an accuracy knob but a
\emph{reliability} knob: a small context makes the model's answer depend heavily
on which rows happened to be drawn, and enlarging the context removes most of
that dependence. The most variable configuration we saw was kc1 at $k{=}16$
(CV $17.8\%$); at $k{=}16$ diabetes AUC over 20 draws ranged from 0.503 to 0.839.

\section{H2: What makes a context good, representativeness or diversity?}

H2 as we first posed it, that a context which preserves the full training
distribution scores higher, turns out to be the wrong reading of a real
correlation. This section builds the case in three steps: the correlation
(observed), a controlled test that overturns its causal interpretation, and a
disentangling experiment that identifies context \emph{diversity/coverage}, not
distribution matching, as the factor that actually tracks accuracy.

Using the 20 random draws per $k$ from H1, we first ask whether the draws that
better preserve the full training distribution score higher. For each draw we
computed two distances to the full training set: the total-variation distance
between class distributions (\texttt{class\_tv}) and the mean absolute
standardized feature-mean shift (\texttt{feat\_shift}). Table~\ref{tab:repr}
reports Pearson $r$, Spearman $\rho$, and linear $R^2$ per dataset, pooled over
$k$.

\begin{table}[h]
\centering
\caption{Correlation between distribution distance and AUC (all draws pooled
over $k$, per dataset). Negative values mean that a context further from the
full distribution scores lower, as hypothesized.}
\label{tab:repr}
\begin{tabular}{lrrrrrr}
\toprule
& \multicolumn{3}{c}{\texttt{class\_tv}} & \multicolumn{3}{c}{\texttt{feat\_shift}} \\
\cmidrule(lr){2-4}\cmidrule(lr){5-7}
Dataset & Pearson & Spearman & $R^2$ & Pearson & Spearman & $R^2$ \\
\midrule
diabetes           & $-0.31$ & $-0.45$ & 0.10 & $-0.52$ & $-0.73$ & 0.27 \\
credit-g           & $-0.53$ & $-0.43$ & 0.28 & $-0.80$ & $-0.83$ & 0.63 \\
vehicle            & $-0.71$ & $-0.78$ & 0.51 & $-0.79$ & $-0.78$ & 0.63 \\
wdbc               & $-0.24$ & $-0.35$ & 0.06 & $-0.39$ & $-0.59$ & 0.15 \\
breast-w           & $-0.18$ & $-0.16$ & 0.03 & $-0.18$ & $-0.27$ & 0.03 \\
blood-transfusion  & $-0.08$ & $-0.18$ & 0.01 & $-0.31$ & $-0.27$ & 0.09 \\
ionosphere         & $-0.26$ & $-0.51$ & 0.07 & $-0.63$ & $-0.84$ & 0.39 \\
sonar              & $-0.53$ & $-0.54$ & 0.28 & $-0.60$ & $-0.66$ & 0.36 \\
glass              & $-0.57$ & $-0.63$ & 0.33 & $-0.56$ & $-0.69$ & 0.32 \\
australian         & $-0.34$ & $-0.26$ & 0.11 & $-0.57$ & $-0.68$ & 0.32 \\
heart-statlog      & $-0.07$ & $-0.23$ & 0.01 & $-0.30$ & $-0.26$ & 0.09 \\
segment            & $-0.77$ & $-0.84$ & 0.59 & $-0.74$ & $-0.82$ & 0.55 \\
waveform-5000      & $-0.71$ & $-0.80$ & 0.51 & $-0.83$ & $-0.92$ & 0.68 \\
spambase           & $-0.39$ & $-0.49$ & 0.15 & $-0.78$ & $-0.89$ & 0.60 \\
kc1                & $-0.08$ & $-0.27$ & 0.01 & $-0.29$ & $-0.38$ & 0.08 \\
\midrule
pooled (z-scored)  & $-0.38$ & $-0.46$ & 0.14 & $-0.55$ & $-0.64$ & 0.31 \\
\bottomrule
\end{tabular}
\end{table}

\begin{table}[h]
\centering
\caption{Mean AUC of the most- vs.\ least-representative third of random draws,
split by \texttt{class\_tv}. The most-representative third wins on all 15
datasets.}
\label{tab:contrast}
\begin{tabular}{lrrr}
\toprule
Dataset & most repr. & least repr. & $\Delta$ \\
\midrule
vehicle           & 0.953 & 0.825 & $+0.127$ \\
sonar             & 0.858 & 0.746 & $+0.112$ \\
waveform-5000     & 0.956 & 0.867 & $+0.088$ \\
glass             & 0.942 & 0.864 & $+0.077$ \\
credit-g          & 0.752 & 0.678 & $+0.074$ \\
diabetes          & 0.851 & 0.791 & $+0.060$ \\
\multicolumn{4}{l}{\textit{\dots (remaining 9 datasets $\Delta$ from $+0.027$ down to $+0.004$)}} \\
breast-w          & 0.994 & 0.990 & $+0.004$ \\
\bottomrule
\end{tabular}
\end{table}

\begin{figure}[h]
\centering
\includegraphics[width=\textwidth]{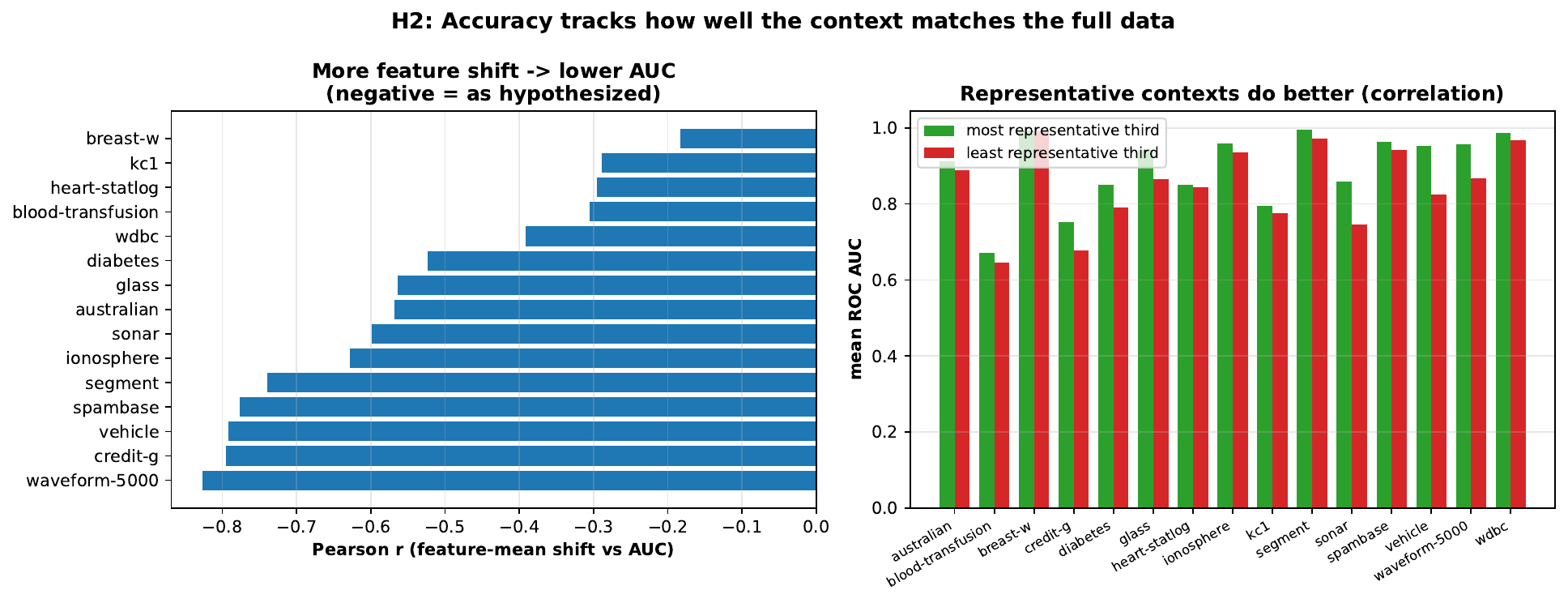}
\caption{The correlational step (later overturned as causal). Left: Pearson
correlation between feature-mean shift and AUC across random draws per dataset
(all negative). Right: mean AUC of the most- vs.\ least-representative third of
random draws. This association is real but confounded with diversity, as the
controlled experiment shows.}
\label{fig:repr}
\end{figure}

\textbf{The correlation.} Across random draws, the direction predicted by H2
holds everywhere: all 15 correlations in Table~\ref{tab:repr} are negative
(further from the full distribution $\rightarrow$ lower AUC), and in
Table~\ref{tab:contrast} the most-representative third of draws beats the
least-representative third on all 15 datasets. \texttt{feat\_shift} is the
stronger predictor (pooled Pearson $-0.55$, $R^2=0.31$), and it is strong on
several datasets, including waveform-5000 ($r=-0.83$, $R^2=0.68$), credit-g
($r=-0.80$), vehicle ($r=-0.79$), and spambase ($r=-0.78$), while weaker near
ceiling (breast-w) or on noisy datasets (kc1, heart-statlog). Taken alone, this looks like support for
H2. It is not, and the rest of this section explains why.

\subsection{A controlled test overturns the causal reading}

Correlation among random draws cannot separate cause from confound. We therefore
\emph{constructed} contexts at target representativeness levels rather than
waiting for random draws to vary: a greedy procedure that adds rows to keep the
running feature-mean as close as possible to the full-data mean (``high''
representativeness, very low \texttt{feat\_shift}), a uniform ``random''
baseline, and a farthest-from-centroid procedure (``low'' representativeness,
high \texttt{feat\_shift}), all with the class ratio matched. These constructed
contexts are designed to isolate causal factors, not to model practical sampling
strategies; their purpose is to move one geometric property at a time. If
representativeness caused accuracy, the low-shift contexts should score highest.

The opposite happens (Table~\ref{tab:controlled}). Forcing low
\texttt{feat\_shift} reduces accuracy substantially, to 0.49 on diabetes and
0.39 on blood-transfusion, far below random, while the high-shift contexts are
as good as or better than random. On most datasets the high-vs-low effect is
large and runs contrary to the original hypothesis (Cohen's $d$ up to $-22.6$,
Cliff's $\delta=-1.0$ on six datasets). So matching the marginal feature means
does not cause good performance; when isolated, it usually degrades it.

\begin{table}[h]
\centering
\caption{Controlled representativeness at $k{=}128$ (15 contexts each, mean ROC
AUC $\pm$ 95\% CI). Forcing low feature-mean shift (``high repr.'') reduces
accuracy substantially; $d$ and $\delta$ are the high-vs-low effect sizes.}
\label{tab:controlled}
\begin{tabular}{lrrrrr}
\toprule
Dataset & high repr. & random & low repr. & Cohen $d$ & Cliff $\delta$ \\
\midrule
diabetes           & $.488{\pm}.027$ & $.859{\pm}.009$ & $.861{\pm}.003$ & $-9.1$ & $-1.00$ \\
credit-g           & $.601{\pm}.024$ & $.765{\pm}.016$ & $.708{\pm}.008$ & $-3.0$ & $-0.97$ \\
vehicle            & $.910{\pm}.007$ & $.945{\pm}.004$ & $.906{\pm}.006$ & $+0.3$ & $+0.26$ \\
wdbc               & $.991{\pm}.001$ & $.992{\pm}.002$ & $.980{\pm}.001$ & $+4.8$ & $+1.00$ \\
breast-w           & $.963{\pm}.001$ & $.994{\pm}.001$ & $.992{\pm}.001$ & $-22.6$ & $-1.00$ \\
blood-transfusion  & $.389{\pm}.016$ & $.694{\pm}.009$ & $.732{\pm}.004$ & $-14.2$ & $-1.00$ \\
ionosphere         & $.966{\pm}.002$ & $.980{\pm}.003$ & $.972{\pm}.003$ & $-1.2$ & $-0.55$ \\
sonar              & $.908{\pm}.004$ & $.893{\pm}.005$ & $.897{\pm}.003$ & $+1.6$ & $+0.76$ \\
glass              & $.953{\pm}.006$ & $.950{\pm}.005$ & $.931{\pm}.002$ & $+2.4$ & $+0.99$ \\
australian         & $.777{\pm}.029$ & $.925{\pm}.007$ & $.929{\pm}.003$ & $-3.5$ & $-1.00$ \\
heart-statlog      & $.866{\pm}.003$ & $.872{\pm}.006$ & $.868{\pm}.004$ & $-0.2$ & $-0.09$ \\
segment            & $.975{\pm}.003$ & $.994{\pm}.001$ & $.969{\pm}.005$ & $+0.7$ & $+0.40$ \\
waveform-5000      & $.699{\pm}.015$ & $.963{\pm}.002$ & $.936{\pm}.007$ & $-10.2$ & $-1.00$ \\
spambase           & $.677{\pm}.036$ & $.977{\pm}.001$ & $.967{\pm}.002$ & $-5.6$ & $-1.00$ \\
kc1                & $.530{\pm}.022$ & $.797{\pm}.016$ & $.661{\pm}.014$ & $-3.4$ & $-0.98$ \\
\bottomrule
\end{tabular}
\end{table}

The pattern is dominant but not universal: on 9 of 15 datasets forcing low shift
reduces accuracy (large negative $d$, $\delta=-1.0$ on 6), with the most dramatic
drops on spambase ($0.68$ vs.\ $0.98$), waveform ($0.70$ vs.\ $0.96$), and the
originals. A handful of near-ceiling datasets (sonar, glass, wdbc) show small
positive effects. The reason for the dominant pattern is a confound: our low-shift
construction also reduces the \emph{diversity} of the context (rows are chosen to
cancel each other's deviations, so they cluster). The correlation in
Table~\ref{tab:repr} arose because, among random draws, a more shifted draw also
tends to be a less diverse one. To find which factor actually matters, we must
vary them independently.

\subsection{Disentangling shift from diversity}

We built six construction methods spanning the (\texttt{feat\_shift},
diversity) plane, ranging from a low-shift/low-diversity clump, to a
low-shift/\emph{high}-diversity set (extremes paired to cancel in the mean), to
high-shift/high-diversity farthest-point sampling. For each context we recorded
\texttt{feat\_shift}, diversity (mean pairwise distance), and coverage
(log-determinant of the context covariance) alongside AUC. We then computed \emph{partial} correlations: the
association of each factor with AUC while holding the other fixed
(Table~\ref{tab:partial}).

\begin{table}[h]
\centering
\caption{Correlation of context geometry with AUC (all six construction methods
pooled, per dataset). ``shift$\mid$div'' is the partial correlation of
feature-mean shift with AUC controlling for diversity, and vice versa; $\beta$
are standardized regression coefficients from AUC $\sim$ shift $+$ diversity.}
\label{tab:partial}
\small
\begin{tabular}{lrrrrrr}
\toprule
Dataset & $r$(shift) & $r$(div) & shift$\mid$div & div$\mid$shift & $\beta_{\text{shift}}$ & $\beta_{\text{div}}$ \\
\midrule
diabetes           & $+0.49$ & $+0.67$ & $+0.34$ & $+0.60$ & $+0.27$ & $+0.57$ \\
credit-g           & $+0.13$ & $+0.35$ & $+0.09$ & $+0.34$ & $+0.08$ & $+0.34$ \\
vehicle            & $-0.62$ & $+0.39$ & $-0.67$ & $+0.49$ & $-0.61$ & $+0.38$ \\
wdbc               & $+0.00$ & $+0.46$ & $-0.37$ & $+0.56$ & $-0.40$ & $+0.69$ \\
breast-w           & $+0.40$ & $+0.57$ & $-0.02$ & $+0.45$ & $-0.03$ & $+0.59$ \\
blood-transfusion  & $+0.26$ & $+0.17$ & $+0.20$ & $+0.03$ & $+0.24$ & $+0.04$ \\
ionosphere         & $-0.65$ & $+0.62$ & $-0.84$ & $+0.83$ & $-0.66$ & $+0.63$ \\
sonar              & $-0.11$ & $-0.07$ & $-0.15$ & $-0.12$ & $-0.17$ & $-0.13$ \\
glass              & $-0.49$ & $+0.02$ & $-0.56$ & $-0.31$ & $-0.66$ & $-0.32$ \\
australian         & $+0.13$ & $-0.33$ & $+0.08$ & $-0.32$ & $+0.08$ & $-0.32$ \\
heart-statlog      & $-0.11$ & $-0.26$ & $-0.15$ & $-0.28$ & $-0.15$ & $-0.28$ \\
segment            & $+0.36$ & $+0.41$ & $-0.12$ & $+0.25$ & $-0.36$ & $+0.75$ \\
waveform-5000      & $+0.27$ & $+0.10$ & $+0.34$ & $+0.24$ & $+0.38$ & $+0.25$ \\
spambase           & $+0.53$ & $+0.48$ & $+0.26$ & $+0.05$ & $+0.46$ & $+0.08$ \\
kc1                & $+0.02$ & $-0.23$ & $+0.50$ & $-0.54$ & $+1.09$ & $-1.20$ \\
\midrule
pooled (z-scored)  & $+0.04$ & $+0.22$ & $-0.01$ & $+0.22$ & $-0.01$ & $+0.23$ \\
\bottomrule
\end{tabular}
\end{table}

\textbf{What we observe.} Diversity is the factor that survives in aggregate.
Pooled across datasets, diversity's partial correlation with AUC (controlling for
shift) is $+0.22$ and its standardized coefficient $+0.23$, whereas feature-mean
shift's partial correlation (controlling for diversity) is $-0.01$ and its
coefficient $-0.01$, essentially zero. The per-dataset picture is mixed rather
than unanimous: diversity's partial correlation is positive on nine of fifteen
datasets (strong on ionosphere $+0.83$, diabetes $+0.60$, wdbc $+0.56$) and
negative on a few (australian, glass, heart-statlog), so the effect is
directionally dominant, not universal. Coverage (log-det of the context
covariance) tracks AUC about as strongly as pairwise diversity on the datasets
where either does (e.g.\ $r=0.72$ on vehicle, $0.69$ on diabetes, $0.67$ on
segment). The controlled construction (Table~\ref{tab:controlled}) remains the
cleaner evidence: the low-shift/high-diversity set scores well while the
low-shift/low-diversity clump scores poorly.

We therefore correct H2. The representativeness--accuracy correlation is real but
\emph{not causal}: it is a shadow of a diversity--accuracy relationship, because
shifted random draws tend also to be narrow ones. What a small-data context needs
appears to be \emph{coverage of the feature space}, not a close match to the
marginal feature distribution. This also explains the cross-method observation
that K-Means and FPS, which are far from distribution-preserving but cover the
space well, match or beat random (H3): they succeed through diversity, the factor
that matters, not through representativeness, the factor that does not. \textbf{Formal inference.} To test this with proper accounting for dataset-level
variation, we fit a linear mixed-effects model \cite{bates2015lme4} over all 945
constructed contexts from the 15 datasets,
\[
\text{AUC} \sim \text{diversity} + \text{shift} + (1\mid\text{dataset}),
\]
with predictors and outcome z-scored within dataset. Diversity is a highly
significant predictor ($\beta=+0.23$, SE $0.03$, $p=3\times10^{-12}$, 95\% CI
$[+0.16,+0.29]$), while feature-mean shift is not significant once diversity is
in the model ($\beta=-0.01$, SE $0.03$, $p=0.71$, 95\% CI $[-0.08,+0.05]$). A
random-slope variant (diversity slope varying by dataset) agrees on the sign and
significance of diversity ($\beta=+0.25$, $p=0.004$); shift is small and, if
anything, negative ($\beta=-0.09$, $p=0.008$), the opposite of what
representativeness would predict. With 15 dataset groups the random-slope
covariance is well-conditioned, unlike the six-dataset pilot. The pooled effect
is smaller than that pilot suggested ($\beta=0.23$ vs.\ $0.47$), which we read as
regression to a more reliable estimate on the broader sample.

We say ``appears to be'' deliberately: the per-dataset partial correlations are
mixed (Table~\ref{tab:partial}), our diversity and coverage measures are
geometric proxies, and the evidence is a controlled association across
constructions, not a randomized single-factor intervention. The claim we stand
behind is the aggregate one, that diversity predicts accuracy and feature-mean
shift does not, and it is backed by the controlled test and the mixed-effects
model.

\begin{figure}[h]
\centering
\includegraphics[width=\textwidth]{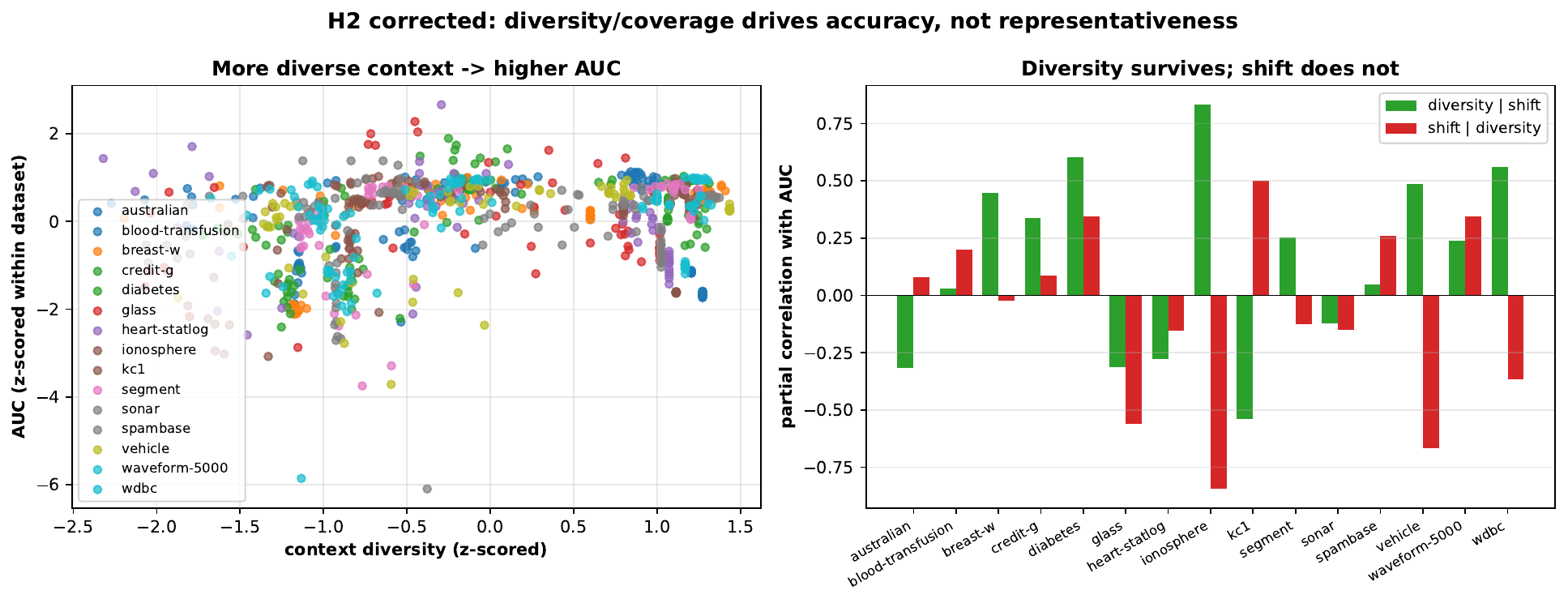}
\caption{H2 corrected. Left: across all constructed contexts, AUC tends to rise
with context diversity (z-scored within dataset). Right: partial correlations per
dataset. Diversity controlling for shift (green) is positive on most datasets,
while shift controlling for diversity (red) is near zero or negative. In
aggregate (pooled and mixed-effects) diversity is the factor that survives.}
\label{fig:disentangle}
\end{figure}

\section{H3: Selection Cost}

At a fixed budget of $k{=}128$ we compared uniform random selection against
K-Means (nearest training point to each centroid, using k-means++ seeding
\cite{arthur2007kmeans}) and farthest-point sampling
(FPS, the classic 2-approximation to $k$-center \cite{gonzalez1985clustering}),
three seeds each. Table~\ref{tab:selection} reports mean AUC and the mean
time to \emph{choose} the prototypes (separate from inference).

\begin{table}[h]
\centering
\caption{Selection methods at $k{=}128$ (mean ROC AUC $\pm$ 95\% CI over three
seeds).}
\label{tab:selection}
\begin{tabular}{lrrr}
\toprule
Dataset & random AUC & K-Means AUC & FPS AUC \\
\midrule
diabetes           & $.869{\pm}.016$ & $.883{\pm}.007$ & $.877{\pm}.004$ \\
credit-g           & $.745{\pm}.057$ & $.755{\pm}.068$ & $.712{\pm}.024$ \\
vehicle            & $.947{\pm}.004$ & $.944{\pm}.014$ & $.947{\pm}.004$ \\
wdbc               & $.994{\pm}.003$ & $.987{\pm}.004$ & $.994{\pm}.001$ \\
breast-w           & $.994{\pm}.004$ & $.995{\pm}.001$ & $.992{\pm}.002$ \\
blood-transfusion  & $.686{\pm}.008$ & $.720{\pm}.009$ & $.725{\pm}.007$ \\
ionosphere         & $.978{\pm}.004$ & $.974{\pm}.007$ & $.960{\pm}.003$ \\
sonar              & $.893{\pm}.004$ & $.895{\pm}.007$ & $.896{\pm}.004$ \\
glass              & $.952{\pm}.005$ & $.938{\pm}.016$ & $.929{\pm}.016$ \\
australian         & $.924{\pm}.012$ & $.934{\pm}.005$ & $.929{\pm}.008$ \\
heart-statlog      & $.855{\pm}.020$ & $.863{\pm}.005$ & $.868{\pm}.007$ \\
segment            & $.995{\pm}.002$ & $.995{\pm}.001$ & $.997{\pm}.001$ \\
waveform-5000      & $.968{\pm}.001$ & $.958{\pm}.006$ & $.968{\pm}.005$ \\
spambase           & $.971{\pm}.005$ & $.975{\pm}.007$ & $.966{\pm}.001$ \\
kc1                & $.800{\pm}.012$ & $.790{\pm}.019$ & $.771{\pm}.018$ \\
\midrule
\multicolumn{4}{l}{\textit{selection time (mean over datasets):}} \\
\multicolumn{4}{l}{random $\approx 0.0003$\,s \quad FPS $\approx 0.04$\,s
\quad K-Means $\approx 0.22$\,s} \\
\bottomrule
\end{tabular}
\end{table}

\begin{figure}[h]
\centering
\includegraphics[width=\textwidth]{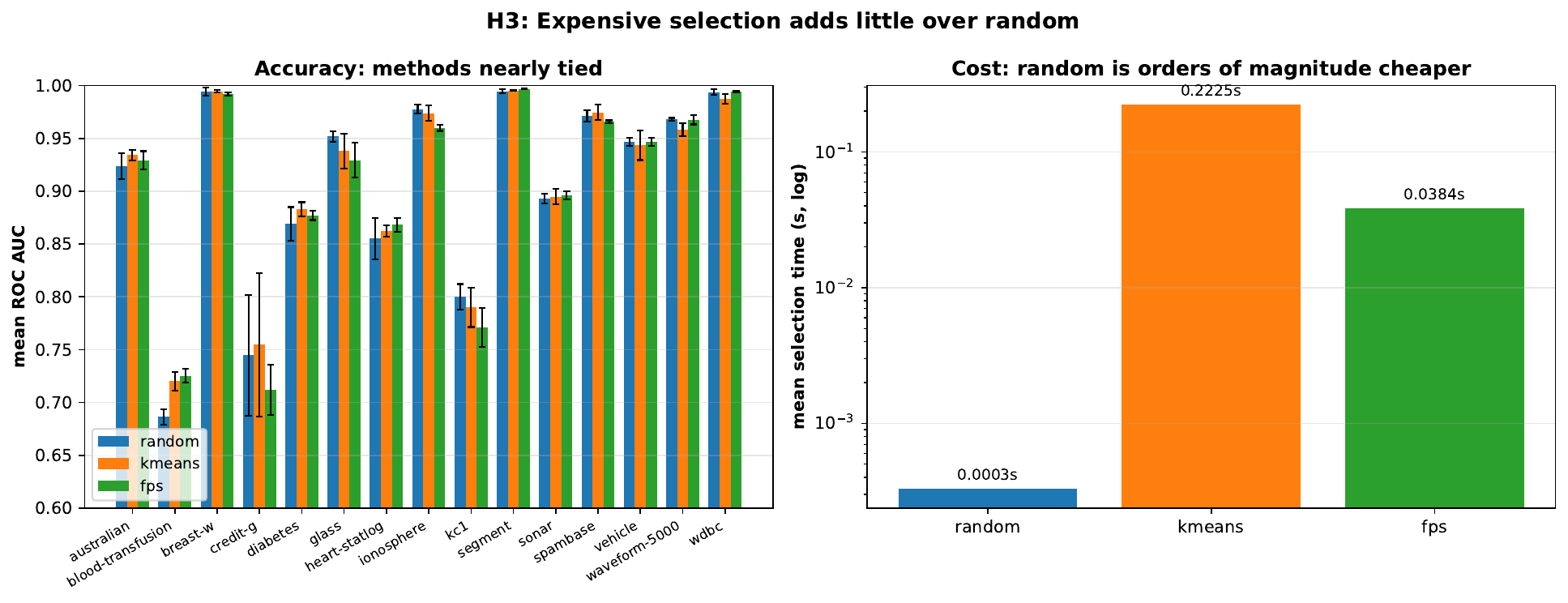}
\caption{Selection methods at $k{=}128$. Accuracy is nearly tied across methods
(left); selection cost differs by two to three orders of magnitude (right).}
\label{fig:selection}
\end{figure}

\textbf{What we observe.} Across all 15 datasets, accuracy differences between the
three methods are small, within about one to three AUC points, and not in a
consistent direction (K-Means marginally ahead on some, FPS or random on others).
The 95\% confidence intervals overlap across methods on most datasets; on
credit-g, where K-Means led random by about one point, the intervals
($0.755{\pm}.068$ vs.\ $0.745{\pm}.057$) overlap heavily, so that gap is not
distinguishable from seed noise. Meanwhile the compute cost differs sharply:
random selection took on the order of $0.0003$\,s versus $\sim$0.22\,s for
K-Means, roughly three orders of magnitude. Given near-equal accuracy, the extra
machinery of K-Means or FPS bought little over drawing rows at random.

This ties back to H2. K-Means and FPS already maximize context diversity, by
spreading prototypes across clusters or across the feature space, which is
exactly the factor H2 identifies as driving accuracy. They perform similarly to
random not despite ignoring distribution matching, but because random already
supplies comparable coverage in expectation. The methods differ mainly in cost,
not in the property that matters.

\section{Supporting: Redundancy and Robustness}

Starting from a random context of $k{=}128$, we either duplicated a fraction of
its rows (adding no new information) or dropped a fraction of them.

\textbf{Redundancy.} Duplicating rows had little effect on most of the 15
datasets: on 13 of them AUC stayed within about 1.5 points up to 70\%
duplication. The clear exception was blood-transfusion, where duplication
steadily hurt ($0.686\rightarrow0.583$), with kc1 and glass showing milder
declines ($\approx-0.015$). So duplicate context rows are mostly harmless when
the signal is strong, but on weak-signal datasets they skew the context and
degrade predictions.

\textbf{Robustness.} Dropping rows degraded accuracy gradually rather than
sharply. Removing 50\% of a 128-row context cost under 1 AUC point on the
stronger datasets (breast-w, diabetes, segment, wdbc) and more on the small or
weak ones (sonar $-0.076$, glass $-0.045$, blood-transfusion $-0.071$). This is
consistent with H1: with fewer rows the context covers less of the space and
predictions get both worse and noisier.

\section{Discussion}

The results fit together into one account of context sampling on small data.
H1 shows that a small context makes TabPFN's prediction unstable, since the
answer depends on which rows were drawn, and that enlarging the context removes
most of that instability while also raising accuracy. H2 asks why some contexts are
better, and the controlled experiments give a sharper answer than the initial
correlation suggested: what matters is the context's \emph{diversity / coverage}
of the feature space, not how closely it matches the training distribution.
Matching the marginals, when forced in isolation, actually hurts (because it
reduces coverage); diversity is the factor that survives once the two are
separated. H3 then follows naturally: uniform random sampling covers the space
well in expectation, so it already captures the factor that matters, and the
more expensive selection methods (also diverse, but no more so in ways that
help) add little accuracy at much higher cost.

A larger $k$ helps in the same terms: more rows cover more of the space and are
less subject to an unlucky, narrow draw. Taken together, on small tabular
datasets the useful things to control are the \emph{size and diversity} of the
context, not its distributional fidelity and not the sophistication of the
selection algorithm. Uniform random sampling at a sufficiently large $k$ is a
strong and cheap default because it delivers both.

\section{Conclusion}

We investigated what makes an in-context training set good for TabPFN on small
tabular datasets. (1) Larger contexts are both more accurate and substantially
more stable, with the coefficient of variation of AUC falling several-fold as
$k$ grows. (2) Accuracy correlates with how representative a random context is,
but a controlled experiment overturns that reading: constructing contexts to
match the feature means \emph{lowers} accuracy (by up to 0.5 AUC), and once we
disentangle the two factors, context diversity/coverage predicts accuracy
(mixed-effects $\beta=+0.23$, $p=3\times10^{-12}$) while feature-mean shift does
not ($\beta=-0.01$, $p=0.71$).
(3) K-Means and farthest-point selection, diverse but not
distribution-preserving, match random within about one AUC point at two to three
orders of magnitude more cost. The practical recommendation is direct: use a
sufficiently large random context and spend no effort on expensive selection. The
scientific takeaway is that what a small-data context needs appears to be
coverage of the feature space, not fidelity to its distribution. Random sampling
succeeds because it provides that coverage for free.

\section{Limitations}

Several limitations bound how far these conclusions should be read. The most
important concerns the strength of the causal claim. Our evidence that diversity,
not representativeness, drives accuracy comes from controlled constructions,
partial correlations, and a mixed-effects model. This is stronger than a raw
correlation, but it is still an association over a designed set of contexts
rather than a randomized intervention on a single factor. The per-dataset partial
correlations are mixed (positive on nine of fifteen datasets), so the diversity
effect is directionally dominant in aggregate but not universal; we say diversity
``appears to be'' the driver accordingly. Diversity and coverage are themselves
measured with geometric proxies (mean pairwise distance, covariance
log-determinant), and a different proxy could shift the details.

The scope of the evidence is also narrow in ways worth stating plainly. We study
fifteen small tabular datasets (roughly 200--5000 rows), so the findings may not
extend to much larger, higher-dimensional, or non-tabular data, and the strength
of H2 on any single dataset depends on its having AUC headroom. We use a single
model (TabPFN v3); whether the same context behavior holds for other in-context
tabular learners is untested. We compare three selection methods (random,
K-Means, FPS), so a method specifically optimized for context quality, for
instance an active-learning-style informativeness criterion
\cite{settles2009active}, could behave differently. Our representativeness
descriptors are two coarse distances (class-TV, feature-mean shift), and richer
divergences (e.g.\ MMD, per-feature KL) might explain more; we leave those to
future work. Finally, all runs used a single consumer GPU (RTX 4060), so the
selection-time comparison reflects that setup and would differ on other hardware.

\section*{Reproducibility}

All experiments are fully reproducible. The complete source code, experiment
scripts, analysis scripts, and figure-generation pipeline are available at:

\begin{center}
\url{https://github.com/mohammed1916/tabmx}
\end{center}

The raw outputs of every experiment are stored under
\texttt{experiments/results/}, including
\texttt{exp14\_\allowbreak context\_\allowbreak sampling\_\allowbreak*.json},
\texttt{exp15\_\allowbreak controlled\_\allowbreak*.json}, and
\texttt{exp16\_\allowbreak disentangle\_\allowbreak*.json}. All tables and
figures in this paper are generated directly from these result files using the
provided analysis scripts
(\texttt{analyze\_\allowbreak exp14.py}, \texttt{analyze\_\allowbreak exp15.py},
\texttt{analyze\_\allowbreak exp16.py},
\texttt{mixed\_\allowbreak effects\_\allowbreak exp16.py}, and
\texttt{generate\_\allowbreak paper2\_\allowbreak figures.py}), ensuring that
every reported number is derived automatically from the logged experiment
outputs.

All datasets are publicly available through OpenML. Dataset identifiers,
random seeds, train/test splits, and experiment configurations are recorded in
the result files. Experiments were performed using TabPFN v3 on an NVIDIA RTX
4060 GPU.


\begin{thebibliography}{15}

\bibitem{hollmann2025tabpfn}
Hollmann, N., Müller, S., Purucker, L., Krishnakumar, A., Körfer, M.,
Hoo, S.~B., Schirrmeister, R.~T., \& Hutter, F. (2025).
Accurate predictions on small data with a tabular foundation model.
\textit{Nature}, 637(8045), 319--326. DOI: 10.1038/s41586-024-08328-6.

\bibitem{muller2022pfn}
Müller, S., Hollmann, N., Pineda Arango, S., Grabocka, J., \& Hutter, F. (2022).
Transformers can do Bayesian inference.
In \textit{International Conference on Learning Representations (ICLR)}.

\bibitem{brown2020gpt3}
Brown, T.~B., Mann, B., Ryder, N., Subbiah, M., et al. (2020).
Language models are few-shot learners.
\textit{Advances in Neural Information Processing Systems (NeurIPS)}, 33,
1877--1901.

\bibitem{dong2024icl}
Dong, Q., Li, L., Dai, D., Zheng, C., et al. (2024).
A survey on in-context learning.
In \textit{Proceedings of the 2024 Conference on Empirical Methods in Natural
Language Processing (EMNLP)}.

\bibitem{garcia2012prototype}
García, S., Derrac, J., Cano, J.~R., \& Herrera, F. (2012).
Prototype selection for nearest neighbor classification: taxonomy and empirical
study.
\textit{IEEE Transactions on Pattern Analysis and Machine Intelligence},
34(3), 417--435.

\bibitem{arthur2007kmeans}
Arthur, D., \& Vassilvitskii, S. (2007).
k-means++: the advantages of careful seeding.
In \textit{Proceedings of the 18th Annual ACM--SIAM Symposium on Discrete
Algorithms (SODA)}, 1027--1035.

\bibitem{gonzalez1985clustering}
Gonzalez, T.~F. (1985).
Clustering to minimize the maximum intercluster distance.
\textit{Theoretical Computer Science}, 38, 293--306.

\bibitem{kulesza2012dpp}
Kulesza, A., \& Taskar, B. (2012).
Determinantal point processes for machine learning.
\textit{Foundations and Trends in Machine Learning}, 5(2--3), 123--286.

\bibitem{feldman2020coresets}
Feldman, D. (2020).
Introduction to core-sets: an updated survey.
\textit{arXiv preprint arXiv:2011.09384}.

\bibitem{sener2018coreset}
Sener, O., \& Savarese, S. (2018).
Active learning for convolutional neural networks: a core-set approach.
In \textit{International Conference on Learning Representations (ICLR)}.

\bibitem{wang2018distillation}
Wang, T., Zhu, J.-Y., Torralba, A., \& Efros, A.~A. (2018).
Dataset distillation.
\textit{arXiv preprint arXiv:1811.10959}.

\bibitem{ghorbani2019datashapley}
Ghorbani, A., \& Zou, J. (2019).
Data Shapley: equitable valuation of data for machine learning.
In \textit{Proceedings of the 36th International Conference on Machine Learning
(ICML)}, 2242--2251.

\bibitem{settles2009active}
Settles, B. (2009).
Active learning literature survey.
Computer Sciences Technical Report 1648, University of Wisconsin--Madison.

\bibitem{bates2015lme4}
Bates, D., Mächler, M., Bolker, B., \& Walker, S. (2015).
Fitting linear mixed-effects models using lme4.
\textit{Journal of Statistical Software}, 67(1), 1--48.

\bibitem{vanschoren2014openml}
Vanschoren, J., van Rijn, J.~N., Bischl, B., \& Torgo, L. (2013).
OpenML: networked science in machine learning.
\textit{ACM SIGKDD Explorations Newsletter}, 15(2), 49--60.

\end{thebibliography}
\end{document}